\documentclass[letterpaper, 10 pt, conference]{ieeeconf}  

\IEEEoverridecommandlockouts                              
\usepackage{comment}
\usepackage{graphicx}

\title{\LARGE \bf
Detection, Recognition and Pose Estimation of Tabletop Objects
}

\author{Sanjuksha Nirgude, Kevin DuCharme, \& Namrita Madhusoodanan 
\thanks{}
}

\begin{document}

\maketitle
\thispagestyle{empty}
\pagestyle{empty}

\begin{abstract}

The problem of cleaning a messy table using Deep Neural Networks is a very interesting problem in both social and industrial robotics. This project focuses on the social application of this technology. A neural network model that is capable of detecting and recognizing common tabletop objects, such as a mug, mouse, or stapler is developed. The model also predicts the angle at which these objects are placed on a table,with respect to some reference. Assuming each object has a fixed intended position and orientation on the tabletop, the orientation of a particular object predicted by the deep learning model can be used to compute the transformation matrix to move the object from its initial position to the intended position. This can be fed to a pick and place robot to carry out the transfer.This paper talks about the deep learning approaches used in this project for object detection and orientation estimation. 

\end{abstract}

\section{INTRODUCTION}

Social robotics, or robots working with humans in social environments, is an exciting field. Robots can be used to do daily chores for humans. Here, we take up one of these daily tasks and use deep learning algorithms to make robots perform a particular task. Most of us would love a robot that can clean up our house, set the dinner table, clear up the mess in a room, or just rearrange a messy table. There can be similar applications of this in industrial environments where robots can be used for the management of the tools and equipment. The idea in this project is to apply deep learning to robotics in such a way that it can organize a messy table in the future, perhaps even customize the table for a particular user.

Our project focuses on detecting and recognizing objects on a tabletop, and estimating the orientation of the object using deep learning techniques. Items in the tabletop have a defined home location such as employed by 5S workplace. With the knowledge of the orientation of the object, the transformation matrix required to transform the object to its home location can be computed. The output can then be utilized to instruct a pick and place robot, serial manipulator, humanoid robot, etc. to carry out the transfer of the item.

\section{LITERATURE REVIEW}

Object detection and recognition along with pose estimation has been a topic of research since many years. Computer vision techniques using hand crafted feature descriptors such as SIFT, SURF etc. in conjunction with RANSAC have been used to understand images, irrespective of the scale, translation or rotation \cite{c11}.  This is extremely useful in many applications where objects need to be recognized in dynamic environments. It also helps in pose estimation, which is essential in the robotics industry to perform grasping and placing operations.

Deep learning is an emerging area of machine learning, that uses many layers to represent multiple layers of abstraction. Neural networks has become a good tool to replace hard coded feature descriptors. Convolutional Neural Networks have proven to be very useful in understanding images and extracting relevant features from them.

Deep learning based pose estimation is an ongoing research problem. It has used various network architectures in the past, such as 2D and 3D Convolutional Neural Networks(CNNs) and autoencoders. Here we describe various techniques used by researchers in the field.

\subsection{Using CNNs for Descriptor Learning}

Wohlhart et all \cite{c9} proposed a method to learn feature descriptors using CNNs. In this method, images of an object are taken in different viewpoints, and multiple objects were present. The CNNs were trained to learn the descriptors in a manner such that the Euclidean Distance between descriptors of different objects remains large, and the Euclidean Distance between descriptors from the same object were representative of the similarity between their poses. Finally, when a new test image is given as an input to the system, a k nearest neighbor algorithm was used to find the descriptor that best describes the object, thus giving its category(class) and pose.

\subsection{Using CNNs to Match Point Clouds to 3D Models}

In the Amazon Picking Challenge 2016 the MIT-Princeton Team, A. Zeng et al \cite{c5}, deployed a system that took an image from 15-18 view points with a depth camera and fed that into a fully convolutional network for 2D object segmentation. This resulted in a 3D point cloud that was then aligned with a pre-scanned 3D model using iterative closest point to obtain a 6D pose.

\subsection{Using CNNs with Posed Datasets}

J. Yu et al \cite{c3} put forth a method of using Max-Pooling CNNs to recognize and estimate the pose of an object. The dataset was comprised of varying images grouped based upon object type and pose. Each object has 12 subclasses broken into 30 degree rotation segments to train as different classifiers. Pose is then determined by which of the subclass classifier is identified. D. Liang et al \cite{c2} expand on this work by separating the network into two Deep Belief Networks, adding a second camera to the second DBN and then joining the last layer together for the classifier. While both approaches demonstrated high accuracy in recognition and pose estimation, it would be difficult to apply this approach in large scale due to the nature of the dataset.

\subsection{Descriptor Regression using a Convolutional Autoencoder}

Inspired by Wohlhart et all \cite{c9},  Kehl et all \cite{c4} proposed a method that could perform 3D object detection and pose estimation under clutter and occlusion. The method uses neural networks along with a local voting based approach. A convolutional autoencoder is trained using random patches from RGB-D images with the aim of creating a model that performs descriptor regression. Finally, when a test image is given as an input, patches are sampled out and passed to the model. The algorithm returns a number of candidate votes which are cast only if their matching score surpasses a certain threshold.

\subsection{Using Spatial Transformer Networks}

Another interesting method uses Spatial Transformer Networks(STNs) \cite{c1}. STNs add spatial transformation capabilities to CNNs. STNs learn to store the knowledge of how to transform each training sample in the weights of its layers, in order to ease the classification process. Transforms include cropping, rotations, scaling and non rigid deformations. STNs are superior to pooling layers in CNNs.The Spatial Transformer module consists of three components:a localization network, a grid generator and a sampler.

\subsection{Estimating pose with distance of object from camera}

The technique of object detection, recognition and its spatial location estimation was used by a group to help blind people see the world using audio \cite{c12}.Their project consists of modules where video is captured using a portable camera, it is then streamed to the server where real-time object detection and recognition is done using YOLO(You Only Look Once).Then the 3D location of the object is estimated using the size and location of the bounding box obtained during detection. For position estimation they defined the default heights of the object and user and hard coded it for 20 classes in the classifier then from the height of the bounding box and the height of the object they estimated the depth. This particular algorithm only calculates the distance of object from the camera and not the orientation of the object. Also, this particular concept was not applied to tabletop objects. Hence, range of positions are large compared to tabletop objects.

\subsection{Deliberative Approach}

The deliberative approaches work by using multi-object pose estimation as a combinatorial search over the space of possible rendered scenes of objects \cite{c13}. Hence it can predict and account for occlusions.They successfully demonstrated object recognition and uncertainty-aware localization in challenging scenes with non-modeled clutter.They extended the deliberative approach of PERCH(Perception via search ) such that it can be applied to cluttered environments without prior models of the objects. It also produces uncertainty estimates for object poses. This approach can be applied to the real-time problem of messy table as it consists of more than one objects in a cluttered environment.  

\section{PROBLEM DESCRIPTION}

The goal of this project is to detect and recognize an object on a tabletop and estimate the orientation of the object with respect to a particular reference frame. Every object in the dataset has a fixed ideal spot on the tabletop, where it should be placed in a certain fixed orientation. The project focuses on building a deep learning model that can detect, recognize, and estimate the orientation of a test input image of an object. Once this information is obtained, the transformation matrix required to place the object in its home location from its current location can be calculated. 

This is quite an interesting problem because the output of such a model can be fed as an input to a pick and place robot. This technology can be applied in numerous fields such as in automated supermarkets, industrial robotics, and personal robots.

The idea can be extended to build personal robots that can organize places in a manner that is customized for a specific user.

\section{DATASET}

This project uses the ‘Tabletop’ dataset, developed by Min Sun et all \cite{c10}. This dataset consists of three categories of  commonly found office desk objects each having 10 object instances: mugs, computer mice, and staplers. The image of each object is captured by a camera in 16 different poses. The images are taken in 8 different angles, at two different heights, H1 and H2. In this project, the orientation of the object refers to the 8 different angles,denoted by A1, A2, A3, A4, A5, A6, A7, and A8. 

The dataset comprised of 80 gray-scaled images of size 640x480 pixels per angle for all three objects (mugs, computer mice, and staplers). The dataset also consists of masks for some of these images. 

\section{DATA PREPROCESSING}
    
\begin{figure}
\centering
\includegraphics[width=\columnwidth]{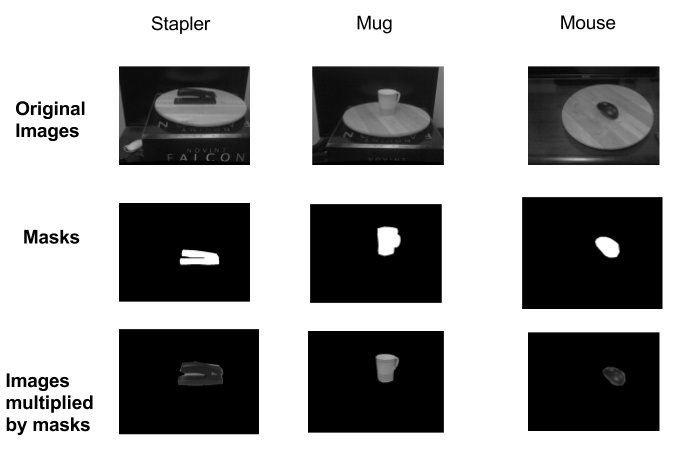}
\caption{Masking of the Images}
\label{fig:dataset} 
\end{figure}

    As all the images in the dataset are gray-scaled, the object of focus on the image cannot be differentiated very well from the background. Additionally, the set of images for the objects under consideration sometimes includes the other objects in the background. For example, images of staplers may include a mouse or mug in the background. To eliminate this problem, the dataset included masks to strip out the background for each image. The information format within the masks was 0 to 255 based, meaning anything to be ignored was 0 and locations to retain was 255. We divided the mask by 255 to transform the image matrix into a logical bit mapping that we could bitwise multiply with the dataset images. This resulted in images that focused only on the object in consideration with the rest of the image being black. However, only a few images had masks, resulting in a very small processed dataset. To counter this problem, the data was augmented. The images multiplied by their masks were shifted horizontally and vertically to produce new images. As the end goal was to predict the orientation of an object, the images were not rotated for augmentation. The resulting dataset comprised of 64,160 images in total, with 8,020 images per angle per object. Fig \ref{fig:dataset} shows a sample of images found in the original dataset, along with their masks, and the images obtained by adding the original images with their masks.

\subsection{Data Parsing}
The images in the dataset were grouped according to the object represented in the image, resulting in three folders: 'mug','mouse', and 'stapler'. Orientation of the objects are defined in the image's file name. A python script was developed to parse the images. Using the OpenCV Library functions, the images were imported and converted to numpy arrays with the orientation also being extracted from the file names. The three numpy arrays representing the images and labels for object type and orientation angle were then pickled for further use.

The dataset was split into two categories. Images that were captured from the 'H1' height viewpoint was used to train the model, while the images that were captured from the  'H2' height viewpoint was used for testing the model.This made sure the testing and training data were mutually exclusive. These heights are included in the filename of each image.

\section{APPROACH}

\begin{figure}
\centering
\includegraphics[width=1.3\columnwidth]{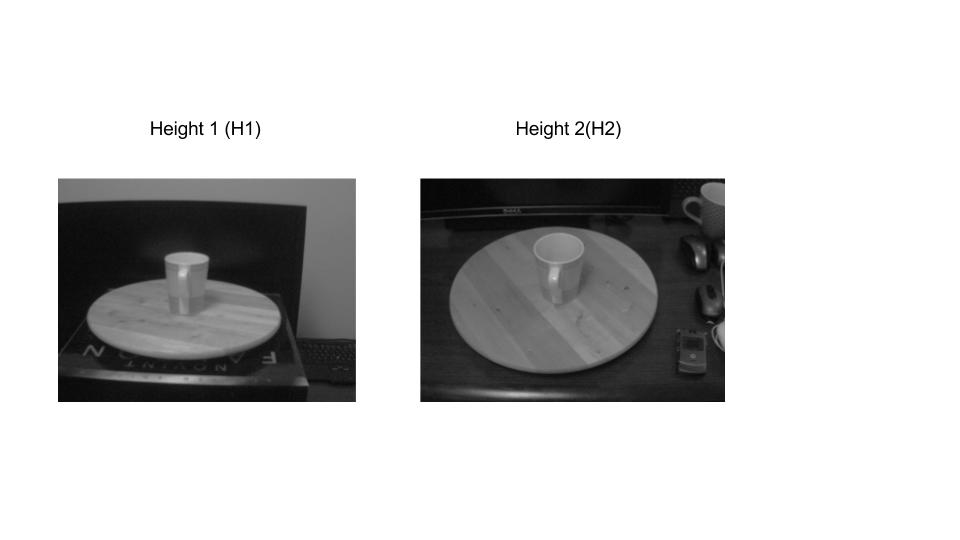}
\caption{Image of mug from two different heights.}
\label{height}
\end{figure}

According to the problem description, we have to first detect and recognize the object represented by a test image. A simple convolutional neural network is used for this purpose. Once this information is obtained, another deep learning model is used to predict the orientation of the object. This is so because a stapler and a mouse at the same angle, say 'A8', appear very different from each other. Therefore, the features that define the orientation of a mouse are very different from those that define the same orientation for a stapler. Hence, three different deep learning models that use convolutional neural networks are trained on datasets of images of mice, staplers, and mug. The models formed are then used to predict the orientation of the object.
As mentioned before, the images captured at height 'H1' are used to train the model and the images captures at height 'H2' are used to test the model. This way we can make sure that the model is tested on images it has not seen during training. Figure \ref{height} depicts the difference between the images of a mug at the angle specified by 'A1' captured at two different heights 'H1' and 'H2'. It can be observed that there is a considerable difference between the two images, thus making this an interesting and challenging problem.  

\section{EXPERIMENTS AND RESULTS}
\subsection{Object Recognition}
Object recognition proved to be a very simple problem. A model comprising of two convolutional layers with 3 x 3 filters and having 64 and 32 activation maps.
This is then followed by a fully connected layer of 300 units, with an output layer behind that. This configuration was able to achieve an accuracy of 98.0 percent. The images were not masked during training or testing, as it was not needed. The model was able to recognize the object of focus without any data preprocessing. For faster training and testing, the sizes of the images were halved, keeping the aspect ratio the same. 

\subsection{Angle Estimation}
To estimate the object orientation angle, the dataset of masked images was employed. Though a lot of experimentation, we finalized on a deep learning model architecture comprising of 5 convolutional layers. Each layer was followed by a max pooling layer stacked with a fully connected layer and an output layer. The first convolutional layer had an output dimensionality of 16, with 5 x 5 filters. The parameters are then halved using a max pooling layer. The next 4 convolutional layers all have 3 x 3 filters with output dimensionality of 32, 64, 70, and 80. All the convolutional layers are activated by the 'ReLU' function, and are followed by a max pooling layer which halves the parameters. After flattening, a fully connected layer of size 300 units was added. To avoid over-fitting, a dropout of 0.3 was added in between the fully connected layer and the final output layer. The output layer was activated by the 'softmax' function. The optimizer used was 'RMSProp'. The models were trained for around 5 epochs and the weights for the models with the highest validation accuracies were saved.

\begin{figure}
\centering
\includegraphics[width=1.3\columnwidth]{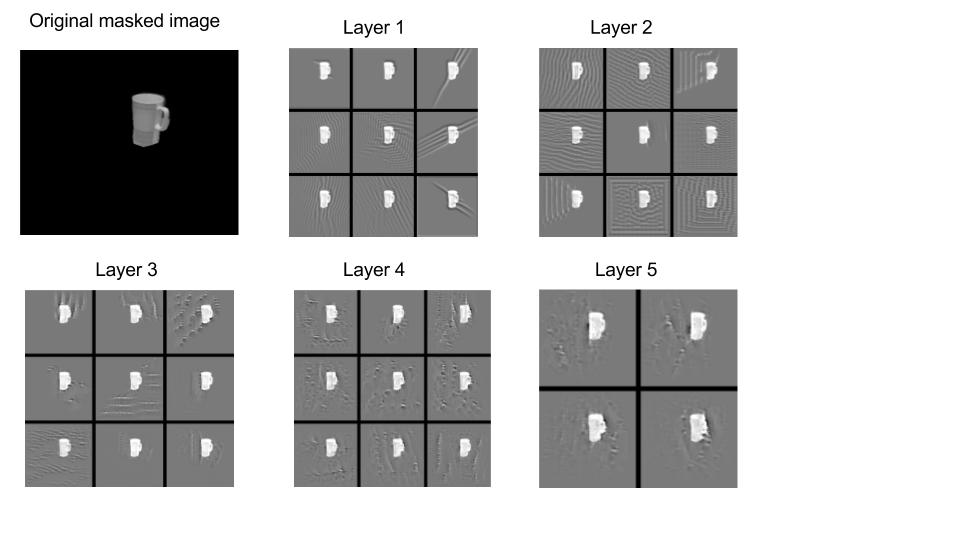}
\caption{Visualization of the CNN layers with the input image is of mug }
\label{visualization}
\end{figure}

The models were all trained using the images captured at height 'H1' and tested on images captured at height 'H2'. The model architecture was identical for all three objects, but consisted of different weights as they were trained separately. During training the validation accuracies achieved were quite high, around 99 percent, for all the three objects. However, when the best performing models were tested using the images captured at height 'H2', the results were different. The model trained for recognizing the orientation of a stapler achieved the best score, around 80 percent accuracy. The model trained for recognizing the orientation of a mug performed second best with an accuracy of 77 percent. The model trained to recognize the orientation of a mouse performed the worst with an accuracy of 55 percent. These results can be attributed to the geometrical features of the three objects. The stapler, being a long and thin object, has the least rotational symmetry. This makes it easier for a neural network model to predict the stapler's orientation. On the other hand, the model does not do a good job while predicting the orientation of the computer mouse due to its much higher degree of rotational symmetry. The filters of each conventional layer were visualized and they are depicted in Figure \ref{visualization}. Here, the image of a mug is given as input. We can see how certain layers extract features that define the orientation of the mug. 

\begin{table} 
\caption{Accuracies}
\label{tab:Accuracy} 
\begin{center}
\begin{tabular}{ |c||c|c|c|}
    \hline
	Object & Stapler & Mug & Mouse \\
    \hline
    \hline
    Training on images captured at H1 & 98.30 & 99.56 & 98.08 \\
    \hline
    Testing on images captured at H2 & 80 & 77 & 55 \\
     \hline
\end{tabular}
\end{center}
\end{table}

Table \ref{tab:Accuracy} shows the validation accuracies of the models while predicting the orientation of the object during training and the final accuracy while testing the models on the test dataset, which comprises of images captured at height 'H2'.

\section{CONCLUSION}
This paper focused on making a deep neural network model to recognize the class and orientation of an object. The objects the network was trained to recognize was mugs, mice, and staplers. Recognizing the object proved to be a very simple task, as the three objects have very different visual features. Object recognition was possible using a simple convolutional neural network.

Determining the orientation of the object proved to be a much more difficult task, which required a lot of data preprocessing. The best performing model for each object category was tested on a dataset that comprised of images taken at a different height. The model for detecting the orientation of staplers performed the best, due to the fact that staplers have a very low degree of rotational symmetry, thus making different orientations more distinct and unique. On the other hand, the model for detecting the orientation of a computer mouse did not perform that well, due to its high degree of rotational symmetry, which is a direct consequence of its oval shape.

\section{FUTURE WORK}
The work demonstrated in this paper can be used to recognize an object from a list of objects and further recognize its orientation with respect to some fixed frame of reference. This information can be extremely useful. For example, the data can be fed to a pick and place robot. The orientation data will determine how the robot must grasp the object. Such robots can be useful in warehouses, automated labs, smart homes, grocery shops,  etc. The robot will be required to compute the transformation matrix that must be applied to the position and orientation of the object to move the object to its intended destination location.

\addtolength{\textheight}{-12cm}  

\end{document}